\documentclass[jair,twoside,11pt,theapa]{article}
\usepackage{jair, theapa, rawfonts}

\usepackage{amsmath,amssymb}
\DeclareMathOperator{\E}{\mathbb{E}}
\usepackage{bbm}
\usepackage{enumitem}
\usepackage{fixltx2e}
\usepackage{mathtools}
\usepackage{multirow}
\usepackage{ctable}
\usepackage[linesnumbered,ruled]{algorithm2e}
\SetEndCharOfAlgoLine{}
\SetKwComment{Comment}{$\triangleright$\ }{}
\usepackage{algpseudocode}

\firstpageno{1}

\begin{document}

\title{Learning Concept Embeddings for Dataless Classification \\via Efficient Bag of Concepts Densification}

\author{\name Walid Shalaby \email wshalaby@uncc.edu \\
       \name Wlodek Zadrozny \email wzadrozn@uncc.edu \\
       \addr Computer Science Department,\\
       University of North Carolina at Charlotte,\\
       Charlotte, NC, 28223 USA}

\researchnote\\
\technicaladdendum

\maketitle

\begin{abstract}
	
	Explicit concept space models have proven efficacy for text representation in many natural language and text mining applications. The idea is to embed textual structures into a semantic space of concepts which captures the main ideas, objects, and the characteristics of these structures. The so called Bag of Concepts (BoC) representation suffers from data sparsity causing low similarity scores between similar texts due to low concept overlap. To address this problem, we propose two neural embedding models to learn continuous concept vectors. Once they are learned, we propose an efficient vector aggregation method to generate fully continuous BoC representations. We evaluate our concept embedding models on three tasks: 1) measuring entity semantic relatedness and ranking where we achieve 1.6\% improvement in correlation scores, 2) dataless concept categorization where we achieve state-of-the-art performance and reduce the categorization error rate by more than 5\% compared to five prior word and entity embedding models, and 3) dataless document classification where our models outperform the sparse BoC representations. In addition, by exploiting our efficient linear time vector aggregation method, we achieve better accuracy scores with much less concept dimensions compared to previous BoC densification methods which operate in polynomial time and require hundreds of dimensions in the BoC representation.

\end{abstract}

\begin{table}[!ht]
	\centering
	\footnotesize
	\begin{tabular}{|l|p{40mm}|p{40mm}|c|c|c|}
		\hline
		\multicolumn{1}{|c|}{\textbf{Category}} & \multicolumn{1}{c|}{\textbf{Top 3 Concepts}} & \multicolumn{1}{c|}{\textbf{Instance Top 3 Concepts}} & \textbf{ESA} & \textbf{CCX} & \textbf{CRX}  \\ \hline
		\multirow{2}{*}{Hockey}                 & 
		- Detroit Red Wings, \newline
		- History of the Detroit Red Wings, \newline
		- History of the NHL on United States television 
		& \underline{Instance (53798)}\newline 
		- History of the Detroit Red Wings,\newline 
		- Detroit Red Wings,\newline 
		- Pittsburgh Penguins& 0.73 & 0.95 & 0.95\\ \cline{3-6}
		&  & \underline{Instance (54551)}\newline 
		- Paul Kariya, \newline 
		- Boston Bruins, \newline 
		- Bobby Orr& 0.0 &0.84  & 0.80\\ \hline
		\multirow{2}{*}{Guns}                 & 
		- Waco siege, \newline 
		- Overview of gun laws by nation, \newline 
		- Gun violence in the United States & \underline{Instance (54387)}\newline 
		- Overview of gun laws by nation, \newline 
		- Waco siege, \newline 
		- Gun politics in the United States& 0.71 & 0.94 & 0.93\\  \cline{3-6}
		&  & \underline{Instance (54477)}\newline 
		- Concealed carry in the United States, \newline 
		- Overview of gun laws by nation, \newline 
		- Gun laws in California& 0.33 & 0.80 & 0.75\\ \hline
	\end{tabular}
	\caption{Top 3 concepts generated using ESA \cite{gabrilovich2007computing} for two 20-newsgroups categories (Hockey and Guns) along with top 3 concepts of sample instances. Using exact match similarity scoring (as in ESA) result in low scores between similar instance and category concept vectors. When using concept embeddings (our models), we obtain relatively higher and more representative similarities.}				
	\label {sparse-representation}
\end{table}

\section{Introduction}
\label{Introduction}

Vector space representation models are used to represent textual structures (words, phrases, and documents) as multidimensional vectors. Typically, those models utilize textual corpora and/or Knowledge Bases (KBs) in order to extract and model real-world knowledge. Once acquired, any given text is represented as a {\it vector} in the semantic space. The goal is thus to accurately place similar structures close to each other in that semantic space, while placing dissimilar ones far apart.

Explicit concept space models are one of these {\it vector-based representations} which are motivated by the idea that, high level cognitive tasks such learning and reasoning are supported by the knowledge we acquire from \emph{concepts}\footnote{A concept is an expression that denotes an idea, event, or an object.} \shortcite{song2015open}. Therefore, such models utilize concept vectors (aka bag-of-concepts (BoC)) as the underlying semantic representation of a given text through a process called \emph{conceptualization}, which is mapping the text into relevant concepts capturing its main ideas, objects, events, and their characteristics. The concept space typically include concepts obtained from KBs such as Wikipedia, Probase \shortcite{wu2012probase}, and others. Once the concept vectors are generated, similarity between two concept vectors can be computed using a suitable similarity/distance measure such as {\it cosine}.

Similar to the traditional Bag-of-Words (BoW) representation, the BoC vector is a multidimensional {\it sparse} vector whose dimensionality is the same as the number of concepts in the employed KB (typically {\it millions}). Consequently, it suffers from \emph{data sparsity} causing low similarity scores between similar texts due to low concept overlap. Formally, given a text snippet $T = \{t_1,t_2,...,t_n\}$ of $n$ terms where $n\ge1$, and a concept space $C = \{c_1,c_2,...,c_N\}$ of size $N$. The BoC vector $\mathbf{v}$ = $\{w_1,w_2,...,w_N\} \in \mathbb{R}^N$ of $T$ is a vector of weights of each concept where each $w_i$ of concept $c_i$ is calculated as in equation 1:
\begin{equation}
w_i = \sum_{j=1}^{n} f(c_i,t_j), \\ 1\le i\le N
\end{equation}
Here $f(c,t)$ is a \emph{scoring function} which indicates the degree of \emph{association} between term $t$ and concept $c$. For example, \shortciteA{gabrilovich2007computing} proposed Explicit Semantic Analysis (ESA) which uses Wikipedia articles as concepts and the TF-IDF score of the terms in these article as the association score. Another scoring function might be the co-occurrence count or Pearson correlation score between $t$ and $c$. As we can notice, only very small subset of the concept space would have nonzero scores with a given term\footnote{Unless the term is very common (e.g., a, the, some...etc) and carry no relevant information.}. Moreover, the BoC vector is generated from the top $n$ concepts which have relatively high association scores with the input terms (typically few hundreds). Thus each text snippet is mapped to a very sparse vector of millions of dimensions having only few hundreds nonzero values leading to the {\it BoC sparsity} problem \shortcite{peng2016event}. 

Typically, the {\it cosine} similarity measure is used compute the similarity between a pair of BoC vectors $\mathbf{u}$ and $\mathbf{v}$. Because the concept vectors are very sparse and for space efficiency, we can rewrite each vector as a vector of tuples $(c_i,w_i)$. Suppose that $\mathbf{u}\!=\!\{(c_{n_1},u_1),\dotsc,(c_{n_{|\mathbf{u}|}},u_{|\mathbf{u}|})\!\}$ and $\mathbf{v}\!=\!\{(c_{m_1},v_1),\dotsc,(c_{m_{|\mathbf{v}|}},v_{|\mathbf{v}|})\!\}$, where $u_i$ and $v_j$ are the corresponding weights of concepts $c_{n_i}$ and $c_{m_j}$ respectively. And $n_i$, $m_j$ are the indices of these concepts in the concept space $C$ such that $1\!\le\!n_i,m_j\!\le\!N$. Then, the similarity score can be written as in equation 2:
\begin{equation}
\begin{multlined}
Sim_{cos}(\mathbf{u},\mathbf{v}) = \frac{\sum_{i=1}^{|\mathbf{u}|}\sum_{j=1}^{|\mathbf{v}|}\mathbbm{1}(n_i\mathord{=}m_j) u_iv_j}{\sqrt{\sum_{i=1}^{|\mathbf{u}|}u_i^2} \sqrt{\sum_{j=1}^{|\mathbf{v}|}v_j^2}}
\end{multlined}
\end{equation}
where $\mathbbm{1}$ is the indicator function which returns 1 if $n_i\mathord{=}m_j$ and 0 otherwise.
Having such sparse representation and using exact match similarity scoring measure, we can expect that two very similar text snippets might have {\it zero similarity} score if they map to \emph{different but very related set of concepts} \shortcite{songunsupervised}. We demonstrate this fact in Table \ref{sparse-representation} (ESA column).


In this paper we utilize \emph{neural-based representations} to overcome the BoC sparsity problem. The basic idea is to {\it map} each concept to a {\it fixed size continuous vector\footnote{We use the terms continuous, dense, distributed vectors interchangeably to refer to real-valued vectors.}}. These vectors can then be used to compute concept-concept similarity and thus overcome the concept mismatch problem.

Our work is also motivated by the success of recent neural-based methods for learning word embeddings in capturing both syntactic and semantic regularities using simple vector arithmetic \cite{mikolov2013efficient,mikolov2013distributed,pennington2014glove}. For example, inferring analogical relationships between words: {\it vec(king)-vec(man)+vec(woman)=vec(queen)}. This indicates that the learned vectors encode meaningful multi-clustering for each word.

However, word vectors suffer from significant limitations. First, each word is assumed to have a \emph{single meaning} regardless of its context and thus is represented by a single vector in the semantic space (e.g., {\it charlotte (city)} vs. {\it charlotte (given name)}). Second, the space contains vectors of single words only. Vectors of multiword expressions (MWEs) are typically obtained by averaging the vectors of individual words. This often produces inaccurate representations especially if the meaning of the MWE is different from the composition of meanings of its individual words (e.g., {\it vec(north carolina)} vs. {\it vec(north)+vec(carolina)}. Additionally, mentions that are used to refer to the same concept would have different embeddings (e.g., {\it u.s., america, usa}), and the model might not be able to place those individual vectors in the same sub-cluster, especially the rare surface forms.

We propose two \emph{neural embedding} models in order to learn continuous concept vectors based on the skip-gram model \shortcite{mikolov2013distributed}. Our first model is the \emph{Concept Raw Context} model (CRX) which utilizes raw concept mentions in a large scale textual KB to jointly learn embeddings of both words and concepts. Our second model is the \emph{Concept-Concept Context} model (CCX) which learns the embeddings of concepts  from their conceptual contexts (i.e., contexts containing surrounding concepts only). 

After learning the concept vectors, we propose an \emph{efficient BoC aggregation} method. We perform \emph{weighted average} of the individual concept vectors to generate fully {\it continuous} BoC representations (CBoC). This aggregation method allows measuring the similarity between pairs of BoC in \emph{linear time} which is more efficient than previous methods that require \emph{quadratic} or at least \emph{log-linear} time if optimized (see equation 2). Our embedding models produce more \emph{representative} similarity scores for BoC containing \emph{different but semantically similar} concepts as shown in Table \ref{sparse-representation} (columns 2-3).

We evaluate our embedding models on three tasks:
\begin{enumerate}[topsep=0pt]
	\itemsep0em
	\item An intrinsic task of measuring entity semantic relatedness and ranking where we achieve 1.6\% improvement in correlation scores.
	\item Dataless concept categorization where we achieve state-of-the-art performance and reduce the categorization error rate by more than 5\% compared to five prior word and entity embedding models.
	\item An extrinsic task of dataless document classification which is a learning \emph{protocol} used to perform text categorization without the need for labeled data to train a classifier \shortcite{chang2008importance}. Experimental results show that we can achieve better accuracy using our efficient BoC densification method compared to the original sparse BoC representation with much less concept dimensions.
\end{enumerate}

The contributions of this paper are fourfold: First, we propose two {\it low cost} concept embedding models which learn concept representations from concept mentions in free-text corpora. Our models require few hours rather than days to train. Second, we show through empirical results the efficacy of the learned concept embeddings in measuring entity semantic relatedness and concept categorization. Our models achieve \emph{state-of-the-art} performance on two concept categorization datasets. Third, we propose simple and efficient vector aggregation method to obtain \emph{fully dense BoC in linear time}. Fourth, we demonstrate through experiments on dataless document classification that we can obtain better accuracy using the dense BoC representation with much less dimensions (few in most cases), reducing the \emph{computational cost} of generating the BoC vector significantly.

The rest of this paper is organized as follows: Section \ref{related} reviews related work; Section \ref{learning} describes our proposed embedding models; Section \ref{applications} introduces the applications of the proposed models; Section \ref{experiments} reports experiments and results; and finally, Section \ref{conclusion} provides discussion and concluding remarks.

\section{Related Work}
\label{related}
\noindent
\textbf{Text Conceptualization:} Humans understand languages through multi-step cognitive processes which involves building rich models of the world and making multi-level generalizations from the input text \shortcite{understanding-short-texts}. One way of automating such generalizations is through text conceptualization. Either by extracting basic level concepts from the input text using concept KBs \shortcite{kim2013context,song2015open}, or mapping the whole input into a concept space that capture its semantics as in ESA \shortcite{gabrilovich2007computing} and MSA \shortcite{shalaby2015measuring}. 

One major line of conceptualization research utilizes semi-structured KBs such as Wikipedia in order to construct the concept space which is defined by all Wikipedia article titles. Such models have proven efficacy for semantic analysis of textual data especially short texts where contextual information is missing or insufficient (see \shortciteA{shalaby2015measuring} for examples). 

Another research direction uses more structured concept KBs such as Probase\footnote{https://concept.research.microsoft.com} 
\shortcite{wu2012probase}. Probase is a probabilistic KB of millions concepts and their relationships (basically is-a). It was created by mining billions of Web pages and search logs of Microsoft's Bing\footnote{https://www.bing.com/} repository using syntactic patterns. The concept KB was then leveraged for text conceptualization to support various text understanding tasks 
such as clustering of Twitter messages and News titles \shortcite{song2011short,song2015open}, \shortcite{song2015open}, search query understanding \shortcite{wang2015query}, short text segmentation \shortcite{wang2014head,hua2015short}, and term similarity \shortcite{li2013computing,kim2013context}. 

Despite its effectiveness, the dependency of Probase on syntactic patterns can be a limitation especially for languages other than English. In addition, we expect augmenting and maintaining these syntactic patterns to be costly and labor intensive. We argue that concept embeddings allow simpler and more efficient representations, simply because similarity scoring between concept vectors can be performed using vector arithmetic. While the Probase hierarchy allows only symbolic matching which still suffers data sparsity. On another hand, we spotted some cases where Probase probabilities were atypical\footnote{$p(\mbox{Arabic coffee}\mid \mbox{beverage})=0$}. This is due to learning concept categories from a limited set of syntactic patterns which does not cover all concept mention patterns. Concept embeddings relax this problem by leveraging all concept mentions in order to learn the embedding vector and therefore might be utilized to curate such atypical Probase assertions.
\smallbreak
\noindent
\textbf{Concept/Entity Embeddings:} Neural embedding models have been proposed to learn distributed representations of concepts/entities\footnote{In this paper we use the terms "concept" and "entity" interchangeably.}. \shortciteA{songunsupervised} proposed using the popular Word2Vec model \shortcite{mikolov2013efficient} to obtain the embeddings of each concept by averaging the embeddings of the concept's individual words. For example, the embeddings of \emph{Microsoft Office} would be obtained by averaging the vector of \emph{Microsoft} and the vector of \emph{Office} obtained from the Word2Vec model. Clearly, this method disregards the fact that the semantics of multiword concepts whose composite meaning 
is different from the semantics of their individual words. 

More robust concept and entity embeddings can be learned from the general knowledge about the concept in encyclopedic KB (e.g., its article) and/or from the structure of a hyperlinked KB (e.g., its link graph). Such concept embedding models were proposed by \shortciteA{hu2015entity,li2016joint}, and \shortciteA{yamada2016joint} who all utilize the skip-gram learning technique \shortcite{mikolov2013distributed}, but differ in how they define the context of the target concept.

\shortciteA{li2016joint} extended the embedding model proposed by \shortciteA{hu2015entity} by jointly learning entity and category embeddings from contexts defined by all other entities in the target entity article as well as its category hierarchy in Wikipedia. This method has the advantage of learning embeddings of both entities and categories jointly. However, defining the entity contexts as pairs of the target entity and all other entities appearing in its corresponding article might introduce noisy contexts, especially for long articles. For example, the Wikipedia article for \emph{"United States"} contains links to \emph{"Kindergarten"}, \emph{"First grade"}, and \emph{"Secondary school"} under the \emph{"Education"} section.

\shortciteA{yamada2016joint} proposed a method based on the skip-gram model to jointly learn embeddings of words and entities using contexts generated from surrounding words of the target entity or word. The authors also proposed incorporating \emph{Wikipedia} link graph by generating contexts from all entities with outgoing link to the target entity to better model entity-entity relatedness. 

Our models also learn word and concept embeddings jointly. Mapping both words and concepts into the same semantic space allows us to easily measure word-word, word-concept, and concept-concept semantic similarities. In addition, our CRX model (described in Section \ref{learning}) extends the context of each word/concept by including nearby concept mentions and not only nearby words. Therefore, we better model local contextual information of concepts and words in \emph{Wikipedia}, treated as a textual KB. During training, we generate word-word, word-concept, concept-word, and concept-concept contexts (cf. equation \ref{conc-embed-eqn3}). In \shortciteA{yamada2016joint} model, concept-concept contexts are generated from \emph{Wikipedia} link graph not from their raw mentions in \emph{Wikipedia} text. In the CCX model, we define concept contexts by all surrounding concepts within a window of fixed size. 

Generating contexts from raw text mentions makes our models scalable and \emph{not restricted to hyperlinked encyclopedic textual corpora} only. This facilitates exploiting other free-text corpora with annotated concept mentions (e.g., news stories, scientific publications, medical guidelines...etc). Moreover, our proposed models are \emph{computationally less costly} than \shortciteA{hu2015entity} and \shortciteA{yamada2016joint} models as they require few hours rather than days to train on similar computing resources.

\smallbreak
\noindent
\textbf{BoC Densification:} Densification of the Bag-of-Concepts (BoC) is the process of converting the sparse BoC into a continuous BoC (CBoC) (aka dense BoC) in order to overcome the BoC sparsity problem. The process requires first mapping each concept into a continuous vector using representation learning. \shortciteA{songunsupervised} proposed three different mechanisms for aligning the concepts at different indices given a sparse BoC pair ($\mathbf{u},\mathbf{v}$) in order to increase their similarity score. 

The \emph{many-to-many} mechanism works by 
averaging all pairwise similarities. 
The \emph{many-to-one} mechanism works by aligning each concept in $\mathbf{u}$ with the most similar concept in $\mathbf{v}$ (i.e., its best match). Clearly, the complexity of these two mechanisms is \emph{quadratic}. The third mechanism is the \emph{one-to-one}. It 
utilizes the Hungarian method in order to find an optimal alignment on a one-to-one basis \shortcite{papadimitriou1982combinatorial}. 
This mechanism performed the best on the task of dataless document classification and was also utilized by \shortciteA{li2016joint}. 

\begin{figure}
	\centering
	\includegraphics[width=15cm,keepaspectratio]{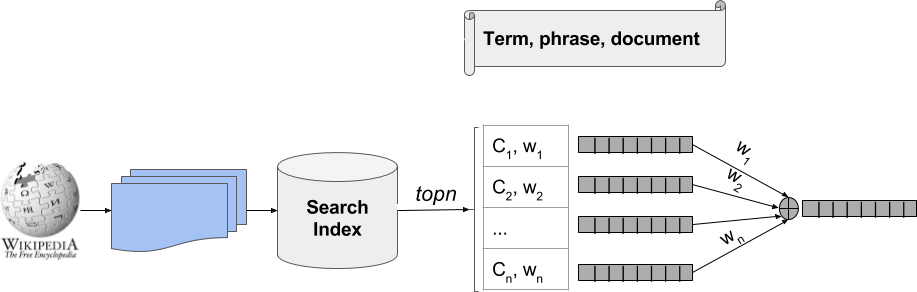}
	\caption[Bag of Concept Densification using Concept Embeddings]{Densification of the bag of concepts vector using weighted average of the learned concept embeddings. The concept space is defined by all Wikipedia article titles. The concept vector is created from the top $n$ hits of searching a \emph{Wikipedia} inverted index with the given text. The weights are the TF-IDF scores from searching \emph{Wikipedia}.}
	\label{embed-boc-dense-arch}
\end{figure}

However, the Hungarian method is a combinatorial optimization algorithm whose complexity is \emph{polynomial}. Our proposed densification mechanism is more efficient than these three mechanisms as its complexity is \emph{linear} with respect to the number of \emph{nonzero} elements in the BoC. Additionally, it is simpler as it does not require tuning a cutoff threshold for the minimum similarity between two aligned concepts as in previous work. Figure \ref{embed-boc-dense-arch} shows a schematic diagram of our efficient densification mechanism applied to a BoC generated from a \emph{Wikipedia} inverted index. We simply perform weighted average of the individual concept vectors in the obtained BoC where concept weights correspond to the TF-IDF scores from searching \emph{Wikipedia}.

\section{Learning Concept Embeddings}
\label{learning}
A main objective of learning concept embeddings is to overcome the inherent problem of \emph{data sparsity} associated with the BoC representation. 
Here we try to learn continuous concept vectors by building upon the skip-gram embedding model \shortcite{mikolov2013distributed}. In the conventional skip-gram model, a set of contexts are generated by sliding a context window of predefined size over sentences of a given text corpus. Vector representation of a target word is learned with the objective to maximize the ability of predicting surrounding words of that target word. 

Formally, given a training corpus of $V$ words $\omega_1, \omega_2, ..., \omega_V$. The skip-gram model aims to maximize the average log probability:
\begin{equation}
\frac{1}{V} \sum_{i=1}^{V}{\sum_{-s \le j \le s,j \ne 0}{\log \ p(\omega_{i+j}|\omega_i)}}
\end{equation}

\noindent where $s$ is the context window size, $\omega_i$ is the target word, and $\omega_{i+j}$ is a surrounding context word. The softmax function is used to estimate the probability $p(\omega_{O}|\omega_I)$ as follows:
\begin{equation}
\label{eq:softmax}
p(\omega_{O}|\omega_I) = \frac{\exp (\mathbf{v}_{\omega_O}^\intercal \mathbf{u}_{\omega_I})}{\sum_{\omega=1}^{W}{\exp(\mathbf{v}_w^\intercal \mathbf{u}_{\omega_I})}}
\end{equation}

\noindent where $\mathbf{u}_{\omega_I}$ and $\mathbf{v}_{\omega_O}$ are the input and output vectors respectively and $W$ is the vocabulary size. \shortciteA{mikolov2013distributed} proposed hierarchical softmax and negative sampling as efficient alternatives to approximate the softmax function which becomes computationally intractable when $W$ becomes huge.

Our approach genuinely learns distributed concept representations by generating concept contexts from \emph{mentions} of those concepts in large encyclopedic KBs such as \emph{Wikipedia}. Utilizing such annotated KBs eliminates the need to manually annotate concept mentions and thus comes at no cost.

\subsection{Concept Raw Context Model (CRX)}
In this model, we jointly learn the embeddings of both words and concepts. First, all concept mentions are identified in the given corpus. Second, contexts are generated for both words and concepts from other surrounding words and other surrounding concepts as well. After generating all the contexts, we use the skip-gram model to jointly learn the embeddings of words and concepts. Formally, given a training corpus of $V$ words $\omega_1, \omega_2, ..., \omega_V$, we iterate over the corpus identifying words and concept mentions and thus generating a sequence of $T$ tokens $t_1, t_2,...t_T$ where $T<V$ (as multiword concepts will be counted as one token). Afterwards we train the a skip-gram model aiming to maximize:
\begin{equation}
\label{conc-embed-eqn3}
\frac{1}{T} \sum_{i=1}^{T}{\sum_{-s \le j \le s,j \ne 0}{\log \ p(t_{i+j}|t_i)}}
\end{equation}

\noindent where as in the conventional skip-gram model, $s$ is the context window size. In this model, $t_i$ is the target token which would be either a word or a concept mention, and $t_{i+j}$ is a surrounding context word or concept mention. 

This model is different from \shortciteA{yamada2016joint}'s anchor context model in three aspects: 1) while generating target concept contexts, we utilize not only surrounding words but also other surrounding concepts, 2) our model aims to maximize $p(t_{i+j}|t_i)$ where $t$ could be a word or a concept, while \shortciteA{yamada2016joint} model maximizes $p(\omega_{i+j}|e_i)$ where $e_i$ is the target concept/entity (see \shortciteA{yamada2016joint} Eq. 6), and 3) in case $t_i$ is a concept, our model captures all the contexts in which it appeared, while \shortciteA{yamada2016joint} model generates for each entity one context of $s$ previous and $s$ next words. We hypothesize that considering both concepts and individual words in the optimization function generates more robust embeddings.

\begin{table*}[]
	\centering
	\footnotesize
	\begin{tabular}{|c|l|l|}
		\hline
		&& \\  [-0.9em]
		\textbf{Sentence} & \multicolumn{1}{c|}{\textbf{CRX Contexts}}                                                     & \multicolumn{1}{c|}{\textbf{CCX Contexts}}                                                        \\ \hline
		&& \\ [-0.9em]
		\begin{tabular}[c]{@{}c@{}}
			\textbf{\emph{Larry Page}} is the \\co-founder of \textbf{\emph{Google}} which is \\headquartered in \textbf{\emph{Menlo Park CA}}
		\end{tabular}                              & \begin{tabular}[c]{@{}l@{}}
			$<$\emph{Larry Page}, co-founder$>$\\
			$<$co-founder, \emph{Google}$>$\\
			$<$\emph{Google}, headquartered$>$\\		
			$<$headquartered, \emph{Menlo Park CA}$>$\\
		\end{tabular}  &
		\begin{tabular}[c]{@{}l@{}}
			$<$\emph{Larry Page}, \emph{Google}$>$\\ $<$\emph{Larry Page}, \emph{Menlo Park CA}$>$\\
			$<$\emph{Google}, \emph{Menlo Park CA}$>$\\ \end{tabular}  
		
		\\ \hline
		&& \\ [-0.9em]
		\begin{tabular}[c]{@{}c@{}}
			\textbf{\emph{Bill Gates}} is the \\co-founder of \textbf{\emph{Microsoft}} which is \\headquartered in \textbf{\emph{Redmond WA}}
		\end{tabular}                              & \begin{tabular}[c]{@{}l@{}}
			$<$\emph{Bill Gates}, co-founder$>$\\
			$<$co-founder, \emph{Microsoft}$>$\\
			$<$\emph{Microsoft}, headquartered$>$\\		
			$<$headquartered, \emph{Redmond WA}$>$\\
		\end{tabular}  &
		\begin{tabular}[c]{@{}l@{}}
			$<$\emph{Bill Gates}, \emph{Microsoft}$>$\\ $<$\emph{Bill Gates}, \emph{Redmond WA}$>$\\
			$<$\emph{Microsoft}, \emph{Redmond WA}$>$\\ 
		\end{tabular}  
		\\ \hline
		&& \\ [-0.9em]
		\begin{tabular}[c]{@{}c@{}}
			\textbf{\emph{Google}} is headquartered in\\ \textbf{\emph{Menlo Park CA}} and was \\co-founded by \textbf{\emph{Larry Page}}
		\end{tabular}                              & \begin{tabular}[c]{@{}l@{}}
			$<$\emph{Google}, headquartered$>$\\		
			$<$headquartered, \emph{Menlo Park CA}$>$\\
			$<$\emph{Menlo Park CA}, co-founded$>$\\
			$<$co-founded, \emph{Larry Page}$>$\\
		\end{tabular}  &
		\begin{tabular}[c]{@{}l@{}}
			$<$\emph{Google}, \emph{Menlo Park CA}$>$\\ $<$\emph{Google}, \emph{Larry Page}$>$\\ $<$\emph{Menlo Park CA}, \emph{Larry Page}$>$
		\end{tabular}  
		\\ \hline
	\end{tabular}
	\caption{Example three sentences along with sample contexts generated from CRX and CCX. Contexts are generated with a context window of length 3.}
	\label{contexts-samples}	
\end{table*}

\subsection{Concept-Concept Context Model (CCX)}
Inspired by the distributional hypothesis \shortcite{harris1954distributional}, in this model, we hypothesize that: \emph{"similar concepts tend to appear in similar conceptual contexts"}. In order to test this hypothesis, we propose learning concept embeddings by training a skip-gram model on contexts generated solely from concept mentions. As in the CRX model, we start by identifying all concept mentions in the given corpus. Then, contexts of target concept are generated from surrounding concepts only. Formally, given a training corpus of $V$ words $\omega_1, \omega_2, ..., \omega_V$. We iterate over the corpus identifying concept mentions and thus generating a sequence of $C$ concept tokens $c_1, c_2,...c_C$ where $C<V$. Afterwards we train the skip-gram model aiming to maximize:
\begin{equation}
\label{embedding-3c-eqn}
\frac{1}{C} \sum_{i=1}^{C}{\sum_{-s \le j \le s,j \ne 0}{\log \ p(c_{i+j}|c_i)}}
\end{equation}

\noindent where $s$ is the context window size, $c_i$ is the target concept, and $c_{i+j}$ is a surrounding concept mention within $s$ mentions. 

This model is different from \shortciteA{li2016joint} and \shortciteA{hu2015entity} as they define the context of a target concept by all the other concepts which have an outgoing link from the concept's corresponding article in \emph{Wikipedia}. 

Formally, given an article about concept $c_t$ containing other concepts ($c_1, c_2, ..., c_n$), \shortciteA{li2016joint,hu2015entity} create context pairs in the form ($c_t$,$c_i$), $1 \le i \le n$. Thus context size is limited to only 2 concepts. 

Clearly, some of these concepts might be irrelevant especially for very long articles which cite hundreds of other concepts. Our CCX model, alternatively, learns concept semantics from surrounding concepts and not only from those that are cited in its article. We also extend the context window beyond pairs of concepts (based on $s$ in equation \ref{embedding-3c-eqn}) allowing more influence to other nearby concepts.

\subsection{CRX vs. CCX}
One of the advantages of the CCX model over the CRX model is its computational efficiency during learning. On the other hand, the CCX model vocabulary is \emph{limited to the corpus concepts} (all \emph{Wikipedia} articles in our case), while the CRX model vocabulary is defined by all \emph{unique concepts+words} in \emph{Wikipedia}. 

Another distinct property of the CCX model is its emphasis on concept-concept \emph{relatedness rather than similarity} (as we will detail more in the experiments section). The CCX model by looking only at surrounding concept mentions while learning, is able to generate contexts containing more diverse but related concepts. One the other hand, the CRX model which jointly learns the embeddings of words and concepts puts more \emph{emphasis on similarity} by leveraging the full contextual information of words and concepts while learning.

To better illustrate this difference, consider a sample of the contexts generated from CRX and CCX in Table \ref{contexts-samples} using a sliding window of length 3. As we can notice, the CRX contexts of \emph{"Google"} and \emph{"Microsoft"} are somewhat similar containing words like \emph{"headquartered"} and \emph{"co-founder"}. This causes the model to learn similar vectors for these two concepts. On the other hand, the CCX contexts of \emph{"Google"} and \emph{"Microsoft"} do not share any similarities\footnote{This is an illustrative example and doesn't imply the two concepts will have totally dissimilar vectors.}, rather we can see that \emph{"Google"} has similar contexts to \emph{"Larry Page"} as both has \emph{"Menlo Park CA"} in their contexts, causing the model to learn similar embeddings for these two related concepts.

\subsection{Training}
We utilize a recent \emph{Wikipedia} dump of August 2016\footnote{http://dumps.wikimedia.org/enwiki/\label{wiki2016}}, which has about 7 million articles. 
We extract articles plain text discarding images and tables. We also discard "References" and "External links" sections (if any). We pruned both articles not under the main namespace and pruned all redirect pages as well. Eventually, our corpus contained about 5 million articles in total. 

We preprocess each article replacing all its references to other \emph{Wikipedia} articles with the their corresponding page IDs. In case any of the references is a title of a redirect page, we use the page ID of the original page to ensure that all concept mentions are normalized. 

Following \shortciteA{mikolov2013distributed}, we utilize negative sampling to approximate the softmax function by replacing every $\log \ p(\omega_{O}|\omega_I)$ term in the softmax function (equation 4) with:
\begin{equation}
\label{eq:-ve-sampling}
\log \sigma(\mathbf{v}_{\omega_O}^\intercal \mathbf{u}_{\omega_I}) + \sum_{s=1}^{k}{\E_{\omega_s\sim P_n(w)} [\log \sigma(-\mathbf{v}_{\omega_s}^\intercal \mathbf{u}_{\omega_I})]}
\end{equation}

\noindent where $k$ is the number of negative samples drawn for each word and $\sigma(x)$ is the sigmoid function ($\frac{1}{1+e^{-x}}$). In the case of the CRX model $\omega_I$ and $\omega_O$ would be replaced with $t_i$ and $t_{i+j}$ respectively. And in the case of the CCX model $\omega_I$ and $\omega_O$ would be replaced with $c_i$ and $c_{i+j}$ respectively.

For both the CRX \& CCX models with use a context window of size 9 and a vector of 500 dimensions. We train the skip-gram model for 10 iterations using 12-core machine with 64GB of RAM. The CRX model took $\sim$15 hours to train for a total of $\sim$12.7 million tokens. The CCX model took $\sim$1.5 hours to train for a total of $\sim$4.5 million concepts.

\subsection{BoC Densification}
As we mentioned in the related work section, the current mechanisms for BoC densification are inefficient as their complexity is at least quadratic with respect to the number of nonzero elements in the BoC vector. Here, we propose simple and efficient vector aggregation method to obtain fully continuous BoC vectors (CBoC) in linear time. Our mechanism works by performing a weighted average of the individual concept vectors in a given BoC. This operation has two advantages. First, it \emph{scales linearly} with the number of nonzero dimensions in the BoC vector. Secondly, it produces a fully dense BoC vector of \emph{fixed size} representing the semantics of the original concepts and \emph{considering their weights}. Formally, given a sparse BoC vector $\mathbf{s}\!=\!\{(c_1,w_1),\dotsc,(c_{|\mathbf{s}|},w_{|\mathbf{s}|})\!\}$, where $w_i$ is weight of concept $c_i$\footnote{The weights are the TF-IDF scores from searching \emph{Wikipedia}.}. We can obtain the dense representation of $\mathbf{s}$ as in equation \ref{eq:dense}:
\begin{equation} 
\label{eq:dense}
\begin{multlined}
\mathbf{s}_{dense} =  \frac{\sum_{i=1}^{|\mathbf{s}|}w_i.\mathbf{u}_{c_i}}{\sum_{i=1}^{|\mathbf{s}|}w_i}
\end{multlined}
\end{equation}
where $\mathbf{u}_{c_i}$ is the vector of concept $c_i$. Once we have this dense BoC vector, we can apply the cosine measure to compute the similarity between a pair of dense BoC vectors. 

As we can notice, this weighted average is done \emph{once} for any given BoC vector. Other mechanisms that rely on concept alignment \cite{songunsupervised}, require \emph{realignment} every time a pair of BoC vectors are compared. Our approach improves the \emph{efficiency} especially in the context of dataless document classification with large number of classes. Using our densification mechanism, 
we apply the weighted average for the BoC of each category and for each instance document once.

Interestingly, our densification mechanism allows us to densify the sparse BoC vector using only the \emph{top few dimensions}. As we will show in the experiments, we can get \emph{near-best} results using these few dimensions compared to densifying with all the dimensions in the original sparse vector. This property reduces the cost of obtaining a BoC vector with a few hundred dimensions in the first place.





\section{Text Conceptualization Applications}
\label{applications}
Concept-based representations have many applications in computational linguistics, information retrieval, and knowledge modeling. Such representations are able to capture the semantics of a given text by either identifying concept mentions in that text, transforming the text into a concept space, or both \shortcite{understanding-short-texts}. Thereafter, many cognitive tasks that require huge background and real-world knowledge are facilitated by leveraging the conceptual representations. We describe some of these tasks in this section, and provide empirical evaluation of our our concept embedding models on such tasks in the next section.

\subsection{Concept/Entity Relatedness}
Entity relatedness has been recently used to model \emph{entity coherence} in many named entity linking and disambiguation systems \shortcite{witten2008effective,milne2008learning,hoffart2012kore,ceccarelli2013learning,huang2015leveraging,hu2015entity,yamada2016joint}. In entity search, \shortciteA{hu2015entity} utilized entity relatedness score to \emph{rank} candidate entities based on their relatedness to the search query entities. Also, entity embeddings have proved more efficient and effective for measuring entity relatedness over traditional relatedness measures which use link analysis.  Formally, given a entity pair ($e_i,e_j$), their relatedness score is evaluated as $rel(e_i,e_j)=Sim(\mathbf{u}_{e_i},\mathbf{u}_{e_j})$, where $Sim$ is a similarity function (e.g., {\it cosine}), and $\mathbf{u}_e$ is the embeddings of entity $e$. 

\subsection{Concept Learning}
Concept learning is a cognitive process which involves classifying a given concept/entity to one or more candidate categories (e.g., {\it milk} as {\it beverage, dairy product, liquid}...etc). This process is also known as {\it concept categorization}\footnote{In this paper, we use concept learning and concept categorization interchangeably} \shortcite{li2016joint}. Automated concept learning gains its importance in many knowledge modeling tasks such as knowledge base {\it construction} (discovering new concepts), {\it completion} (inferring new relationships between concepts), and {\it curation} (removing noisy or assessing weak relationships). Similar to \shortciteA{li2016joint}, we assign a given concept to a target category using Rocchio classification \shortcite{rocchio1971relevance}, where the centroid of each category is set to the category's corresponding embedding vector. Formally, given a set of $n$ candidate concept categories $G = \{g_1, ..., g_n\}$, a sample concept $c$, an embedding function $f$, and a similarity function $Sim$, then $c_i$ is assigned to category $g_*$ such that $g_* = arg\ \mbox{max}_i\ Sim(f(g_i),f(c))$. Here, the embedding function $f$ would always map the given concept to its vector.

\subsection{Dataless Classification}
\shortciteA{chang2008importance} proposed dataless document classification as a learning \emph{protocol} to perform text categorization without the need for labeled data to train a classifier. Given only label names and few descriptive keywords of each label, classification is performed \emph{on the fly} by mapping each label into a BoC representation using ESA \shortcite{gabrilovich2007computing}. Likewise, each data instance is mapped into the same BoC semantic space and assigned to the most similar label using a proper similarity measure such as \emph{cosine}. Formally, given a set of $n$ labels $L = \{l_1, ..., l_n\}$, a text document $d$, a BoC mapping model $f$, and a similarity function $Sim$. First we conceptualize the each $l_i$ and the document $d$ by applying $f$ on them, which will produce sparse BoC vectors $s_{l_i}$ and $s_d$ respectively. Then we densify the vectors as in equation \ref{eq:dense} producing $s_{dense_{{l_i}}}$ and $s_{dense_d}$ respectively. Finally $d$ is assigned to label $l_*$ such that $l_* = arg\ \mbox{max}_i\ Sim(s_{dense_{{l_i}}},s_{dense_d})$.

\begin{algorithm}[t]
	\scriptsize
	\caption{Classification + Bootstrapping}\label{alg1}
	\KwIn{
		$\mathbf{U}\!=\!\{ (l_1,\mathbf{u}_{l_1}),...,(l_n,\mathbf{u}_{l_n})$\}: labels + embeddings \newline
		$\mathbf{D}\!=\!\{ (d_1,\mathbf{v}_{d_1}),...,(d_m,\mathbf{v}_{d_m})$\}: instances + embeddings \newline
		N: number of bootstrap instances
	}
	\KwResult{
		$\mathbf{L}\!=\!\{...,(d_i,l_j),...\}$: label assignment for each instance
	}
	\Repeat{$\mathbf{D} = \phi$ \Comment{no more instances to classify}}{
		$\mathbf{candidates}\gets\{ l_1:\phi,...,l_n:\phi$\}\\
		\ForEach{$(d,\mathbf{v}_{d})\in\mathbf{D}$}{
			$d_{max\_sim} = 0$\\
			$d_{max\_label} = null$\\
			\ForEach{$(l,\mathbf{u}_{l})\in\mathbf{U}$}{
				$sim_{l} = Sim(\mathbf{v}_{d},\mathbf{u}_{l})$\\
				\If{$sim_{l}>d_{max\_sim}$}
				{
					$d_{max\_sim} = sim_{l}$\\
					$d_{max\_lebel} = l$					
				}
			}
			add ($d,d_{max\_sim}$) to $\mathbf{candidates}[l]$\\
		}
		\ForEach{$(l,\mathbf{candidates}_{l})\in\mathbf{candidates}.items$}{
			\Repeat{N highest scored instances added} {
				$score_{max} = 0$\\
				$d_{max} = null$\\
				\ForEach{$(d,score_{d})\in\mathbf{candidates}_{l}$}{
					\If{$score_{d}>score_{max}$}{
						$score_{max} = score_{d}$\\
						$d_{max} = d$ \Comment{most similar instance so far}
					}
				}
				add $(d_{max},l)$ to $\mathbf{L}$\Comment{assign class label}\\
				$\mathbf{u}_{l} \gets \mathbf{u}_{l}  + \mathbf{v}_{d}$\Comment{bootstrap label embedding}\\
				remove $d$ from $\mathbf{candidates}_{l}$\\
				remove $d$ from $\mathbf{D}$\\
			}
		}
	}
	
\end{algorithm}

In the context of dataless classification, \shortciteA{chang2008importance} and \shortciteA{song2014dataless} used bootstrapping in order to improve the classification performance without the need for labeled data. The basic idea is to start from target labels as the initial training samples, train a classifier, and \emph{iteratively} add to the training data those samples which the classifier is \emph{most confident} until no more samples to be classified. The results of dataless classification with bootstrapping were competitive to supervised classification with many training examples. 

In this paper, we extend the use of bootstrapping to the concept learning task as well. In concept learning we start with the vectors of target category concepts as a prototype view upon which categorization decisions are made (e.g., {\it vec(bird), vec(mammal)...etc}). We leverage bootstrapping by \emph{iteratively updating} this prototype view with the vectors of concept instances we are most confident. For example, if \emph{"deer"} is closest to \emph{"mammal"} than any other instance in the dataset, then we update the definition of {\it "mammal"} by performing {\it vec(mammal)+=vec(deer)}, and repeat the same operation for other categories as well. This way, we \emph{adapt} the initial prototype view to better match the specifics of the given data. Although bootstrapping is a time consuming process, we argue that, using dense vectors for representing concepts makes bootstrapping more appealing. As updating the category vector with an instance vector could be performed through optimized vector arithmetic which is available in most modern machines. Algorithm \ref{alg1} presents the pseudocode for performing  dataless classification and concept categorization with bootstrapping. In our implementation, we bootstrap the category vector with vectors of the most similar $\mathbf{N}$ instances at a time. Another implementation option might be defining a threshold and bootstrapping using vectors of $\mathbf{N}$ instances if their similarity score exceed that threshold. In the experiments, we set $\mathbf{N}\!=\!1$.

%

\begin{table*}[]
	\centering
	{\setlength{\tabcolsep}{0.5em}
		\begin{tabular}{c|cccccc} 
			\hline
			\\ [-0.9em]
			Method                   & IT Companies & Celebrities & TV Series & Video Games  & Chuck Norris & All          \\ [0.1em] \hline 
			&&&&&& \\ [-0.9em]
			WLM  & 0.721&0.667&0.628&0.431&0.571& 0.610    \\ [0.2em]
			CombIC  &0.644 &0.690&0.643&0.532&0.558& 0.624     \\ [0.2em]
			ExRel &0.727&0.643&0.633&0.519&0.477&0.630    \\ [0.2em]
			KORE  &0.759&\textbf{0.715}&0.599&\textbf{0.760}&0.498& 0.698   \\ [0.2em] \hline
			CRX  &0.644 &0.592&0.511&0.641&0.495& 0.586    \\ [0.2em]
			CCX  &\textbf{0.788} &\underline{0.694}&\textbf{0.696}&\underline{0.708}&\textbf{0.573}& \textbf{0.714} \\ 
			\hline
	\end{tabular}}
	\caption{Evaluation of concept embeddings for measuring entity semantic relatedness using Spearman rank-order correlation (\textit{$\rho$}). Overall, the CCX model gives the best results outperforming all other models. It comes 1$^{st}$ on 3 categories (bold), and 2$^{nd}$ on the other two (underlined).}		
	\label{entity-relatedness-tbl}
\end{table*}

\section{Experiments}
\label{experiments}
\subsection{Entity Semantic Relatedness}
We evaluate the "goodness" of our concept embeddings on measuring entity semantic relatedness as an intrinsic evaluation. 

\subsubsection{Dataset}
We use the KORE dataset created by \shortciteA{hoffart2012kore}. It consists of 21 main entities from four domains: IT companies, Hollywood celebrities, video games, and television series. For each of these entities, 20 other candidate entities were selected and manually ranked based on their relatedness score based on human judgements. As in previous studies, we report the Spearman rank-order correlation (\textit{$\rho$}) \cite{zwillinger1999crc} which assesses how the automated ranking of candidate entities based on their relatedness score matches the ranking we obtain from human judgements.

\subsubsection{Compared Systems}
We compare our models with four previous methods:
\begin{enumerate}[topsep=0pt]
	\itemsep0em
	\item \textbf{\emph{KORE}} \cite{hoffart2012kore} which measure entity relatedness by firstly representing entities as sets of weighted keyphrases and then computing relatedness using different measures such as keyphrase vector cosine similarity and keyphrase overlap relatedness.
	\item \textbf{\emph{WLM}} introduced by \shortciteA{witten2008effective} who proposed a Wikipedia Link-based Measure (WLM) as a simple mechanism for modeling the semantic relatedness between \emph{Wikipedia} entities. The authors utilized \emph{Wikipedia} link structure under the assumption that related entities would have similar incoming links.
	\item \textbf{\emph{Exclusivity-based Relatedness (ExRel)}} introduced by \shortciteA{hulpucs2015path} who proposed this measure under the assumption that not all instances of a given relation type should be equally weighted. Specifically, the authors hypothesized that the relatedness score between two concepts should be higher if each of them is related through the same relation type to fewer other concepts in the employed KB link graph.
	\item \textbf{\emph{Combined Information Content (CombIC)}} introduced by \shortciteA{schuhmacher2014knowledge} who compute the relatedness score using a graph edit distance measure on the \emph{DBpedia KB}.
\end{enumerate}

\begin{table*}[]
	\centering
	\footnotesize
	\begin{tabular}{|c|l|l|l|}
		\hline
		&&& \\  [-0.9em]
		\textbf{Entity} & \multicolumn{1}{c|}{\textbf{CRX}}                                                     & \multicolumn{1}{c|}{\textbf{CCX}} & \multicolumn{1}{c|}{\textbf{Ground Truth (GT)}}                                                        \\ \hline
		&&& \\ [-0.9em]
		Google                           & \begin{tabular}[c]{@{}l@{}}
			Yahoo!	(9) \\
			Apple Inc.	(12) \\
			Bing (search engine) (7)
			
		\end{tabular}    & 
		\begin{tabular}[c]{@{}l@{}}
			\underline{Larry Page}	(1) \\
			\underline{Sergey Brin} (2) \\
			Yahoo! (9)
		\end{tabular} &
		\begin{tabular}[c]{@{}l@{}}
			Larry Page	\\
			Sergey Brin	\\
			Google Maps				
		\end{tabular}  
		\\ \hline
		&&& \\ [-0.9em]
		\begin{tabular}[c]{@{}c@{}}
			Leonardo\\ DiCaprio
		\end{tabular}                            & \begin{tabular}[c]{@{}l@{}}
			Kate Winslet (4)	\\
			Steven Spielberg (9)	\\
			Tobey Maguire (7)	
		\end{tabular}    & 
		\begin{tabular}[c]{@{}l@{}}
			Tobey Maguire (7)	\\
			Kate Winslet (4)	\\
			\underline{Titanic (1997 film)}	(2)
		\end{tabular} &
		\begin{tabular}[c]{@{}l@{}}
			Inception (film)	\\
			Titanic (1997 film)	\\
			Frank Abagnale
		\end{tabular}  
		\\ \hline
		&&& \\ [-0.9em]
		Mad Men                           & \begin{tabular}[c]{@{}l@{}}
			The Sopranos (15) \\
			\underline{Matthew Weiner} (1)  \\
			\underline{Jon Hamm} (2)
		\end{tabular}    & 
		\begin{tabular}[c]{@{}l@{}}
			\underline{Matthew Weiner} (1) \\	
			\underline{Jon Hamm} (2)	\\
			Todd London	(4)
		\end{tabular} &
		\begin{tabular}[c]{@{}l@{}}
			Matthew Weiner	\\
			Jon Hamm	\\
			Alan Taylor (director)	
		\end{tabular}  
		\\ \hline
		&&& \\ [-0.9em]
		\begin{tabular}[c]{@{}c@{}}
			Guitar Hero \\(video game)
		\end{tabular}                              & \begin{tabular}[c]{@{}l@{}}
			Frequency (video game) (10)	\\
			Rock Band (video game) (6)	\\
			\underline{Harmonix Music Systems} (1)	
		\end{tabular}                                   & \begin{tabular}[c]{@{}l@{}}
			\underline{Harmonix Music Systems} (1)	\\
			\underline{WaveGroup Sound}	(3) \\
			\underline{RedOctane} (1)	
		\end{tabular}  &                     
		\begin{tabular}[c]{@{}l@{}}
			Harmonix Music Systems	\\
			RedOctane	\\
			WaveGroup Sound
		\end{tabular}                               \\ \hline
		
	\end{tabular}
	\caption{Top-3 rated entities from CRX \& CCX models on sample entities from the 4 domains compared to the ground truth. We can notice high agreement between CCX model ranks and the ground truth ranks (in brackets). The CRX model top rated entities has lower ranks than ground truth ranks causing relatively low correlation scores.}
	\label{entity-relatedness-samples}	
\end{table*}

\subsubsection{Results}
Table \ref{entity-relatedness-tbl} shows the Spearman (\textit{$\rho$}) correlation scores of the CRX and CCX model compared to previous models. As we can notice the CCX model achieves the best overall performance on the five domains combined exceeding its successor KORE by 1.6\%. The CRX model on the other hand came last on this task.

In order to better understand these results, we looked at rankings of individual entities from each domain to see how they compare to the ground truth. Table \ref{entity-relatedness-samples} shows the top-3 rated entities from each model on sample entities from the four domains. As we can notice, the ground truth assigns high rank to related rather than similar entities. For example, relatedness of \emph{"Google"} to \emph{"Larry Page"} is ranked 1st, while to \emph{"Yahoo!"} is ranked 9\textsuperscript{th}, and to \emph{"Apple Inc."} is ranked 12\textsuperscript{th}. As the CCX model emphasizes semantic relatedness over similarity, it has high overlap in the top-3 entities with the ground truth (underlined entities). On the other hand, the CRX model predictions are actually meaningful when it comes to functional and topical similarity. As we can notice, it assigns high ranks of \emph{"Google"} to other companies ("Yahoo!", "Apple Inc."), of "Leonardo DiCaprio" to other celebrities (\emph{"Tobey Maguire"}), and \emph{"Mad Men"} to other TV series (\emph{"The Sopranos"}), and of \emph{"Guitar Hero"} to other video games (\emph{"Frequency"}, \emph{"Rock Band"}). However, all these highly ranked entities by CRX have relatively low rankings in the ground truth (given in brackets). This caused the correlation score to be much lower than what we obtained from the CCX model. 

The results indicate that, the CCX model could be more appropriate in applications where relatedness and topical diversity are more desired than topical and functional coherence where the CRX model would be more appealing.

\begin{table*}[]
	\centering
	{\setlength{\tabcolsep}{0.5em}
		\begin{tabular}{c|cccc} 
			\hline
			\\ [-0.9em]
			\underline{Dataset/Instances}                   & Battig & DOTA-single & DOTA-mult & DOTA-all          \\
			Method & (83) & (300) & (150) & (450)          \\ \hline 
			WE$_{Senna}$ & 0.44  & 0.52 & 0.32 & 0.45 \\ [0.1em] \hline
			WE$_{Mikolov}$  & 0.74  & 0.72 & 0.67 & 0.72 \\ [0.1em] \hline
			TransE\textsubscript{1}  & 0.66  & 0.72 & 0.69 & 0.71 \\ [0.1em] \hline
			TransE\textsubscript{2}  & 0.75  & 0.80 & 0.77 & 0.79 \\ [0.1em] \hline
			TransE\textsubscript{3}  & 0.46  & 0.55 & 0.52 & 0.54 \\ [0.1em] \hline
			CE  & 0.79  & 0.89 & 0.85 & 0.88 \\ [0.1em] \hline
			HCE  & 0.87  & 0.93 & 0.91 & 0.92 \\ [0.1em] \hline
			CCX  & 0.72  & 0.90 & 0.80 & 0.87 \\ [0.1em]
			+boostrap  & 0.81  & 0.91 & 0.85 & 0.87 \\ [0.1em] \hline 
			CRX  & 0.83  & 0.91 & 0.88 & 0.90 \\ [0.1em]
			+bootstrap  & \textbf{0.89}  & \textbf{0.98} & \textbf{0.95} & \textbf{0.97} \\ 
			\hline
	\end{tabular}}
	\caption{Accuracy of concept categorization. The CRX model with bootstrapping gives the best results outperforming all other models.}		
	\label{concept-categorization}
\end{table*}

\subsection{Concept Categorization}
This task can be viewed as both intrinsic and extrinsic. It is intrinsic because a \emph{good} embedding model would generate clusters of concepts belonging to the same category, and optimally place the category vector at the center of its instances vectors. On another hand, it is extrinsic as the embedding model could be used to generate a concept KB of is-a relationships with confidence scores, similar to \emph{Probase} \shortcite{wu2012probase}. The model could even be used to curate and/or assert the facts in \emph{Probase}.

\subsubsection{Datasets}
As in \shortciteA{li2016joint}, we utilize two benchmark datasets: 1) Battig test \cite{baroni2010distributional}, which contains 83 single word concepts (e.g., {\it cat, tuna, spoon..etc}) belonging to 10 categories (e.g., {\it mammal, fish, kitchenware..etc}), and 2) DOTA which was created by \shortciteA{li2016joint} from \emph{Wikipedia} article titles (entities) and category names (categories). DOTA contains 300 single-word concepts (DOTA-single) (e.g., {\it coffee, football, semantics..etc}), and (150) multiword concepts (DOTA-mult)  (e.g., {\it masala chai, table tennis, noun phrase...etc}). Both belong to 15 categories (e.g., {\it beverage, sport, linguistics..etc}). Performance is measured in terms of the ability of the system to assign concept instances to their correct categories.

\subsubsection{Compared Systems}
We compare our models to various word, entity, and category embedding methods as described in \shortciteA{li2016joint} including:
\begin{enumerate}[topsep=0pt]
	\itemsep0em
	\item \textbf{\emph{Word embeddings}:} \shortciteA{collobert2011natural} model (WE$_{Senna}$) trained on \emph{Wikipedia}. Here vectors of multiword concepts are obtained by averaging their individual word vectors.
	\item \textbf{\emph{MWEs embeddings}:} \shortciteA{mikolov2013distributed} model (WE$_{Mikolov}$) trained on \emph{Wikipedia}. This model jointly learns single and multiword embeddings where MWEs are identified using corpus statistics.
	\item \textbf{\emph{Entity-category embeddings}:} which include \shortciteA{bordes2013translating} embedding model (TransE). This model utilizes relational data between entities in a KB as triplets in the form (entity,relation,entity) to generate representations of both entities and relationships. \shortciteA{li2016joint} implemented three variants of this model (TransE$_1$, TransE$_2$, TransE$_3$) to generate representations for entities and categories jointly. Two other models introduced by \shortciteA{li2016joint} are CE and HCE. CE generates embeddings for concepts and categories using category information of \emph{Wikipedia} articles. HCE extends CE by incorporating \emph{Wikipedia}'s category hierarchy while training the model to generate concept and category vectors.
\end{enumerate}

\subsubsection{Results}
We report the accuracy scores of concept categorization\footnote{From a multi-class classification perspective, the accuracy scores would be equivalent to the clustering purity score as reported in \shortciteA{li2016joint}.} in Table \ref{concept-categorization}. Accuracy is calculated by dividing the number of correctly classified concepts by the total number of concepts in the given dataset. Scores of all other methods are obtained from \shortciteA{li2016joint}. As we can see in Table \ref{concept-categorization}, the CRX model comes second after the HCE on all datasets. While the CCX model performance is much less than CRX. With bootstrapping, the CCX model performance improves on both datasets. CRX with bootstrapping outperforms all other models by significant percentages. These results show that learning concept embeddings from concept mentions is actually different from training the skip-gram model on phrases or multiword expressions. This is clear from the significant performance gains we get from the CRX and CCX models compared to WE$_{Mikolov}$ which was trained using skip-gram on phrases. Additionally, the results demonstrate the efficacy of our models which simply learn concept embeddings from concept mentions in free-text corpus compared to the more complex models which require category or relational information such as TransE, CE, and HCE.
	
\subsection{Dataless Classification}
In this experiment, we evaluate the effectiveness of our concept embedding models on the dataless document classification task as an extrinsic evaluation. We demonstrate through empirical results the efficiency and effectiveness of our proposed BoC densification scheme which helps obtaining better classification results compared to the original sparse BoC representation.

\subsubsection{Dataset}
We use the 20-newsgroups dataset (20NG) \shortcite{lang1995newsweeder} which is commonly used for benchmarking text classification algorithms. The dataset contains 20 categories each has $\sim$1000 news posts. We obtained the BoC representations using ESA from \shortciteA{song2014dataless} who utilized a Wikipedia index containing pages with 100+ words and 5+ outgoing links to create ESA mappings of 500 dimensions for both the categories and news posts of the 20NG. We designed two types of classification tasks: 1) fine-grained classification involving closely related classes such as \emph{Hockey} vs. \emph{Baseball}, \emph{Autos} vs. \emph{Motorcycles}, and \emph{Guns} vs. \emph{Mideast} vs. \emph{Misc}, and 2) coarse-grained classification involving top-level categories such as \emph{Sport} vs. \emph{Politics} and \emph{Sport} vs. \emph{Religion}. The top-level categories are created by combining instances of the fine-grained categories which are shown in Table \ref{category-mappings}.

\begin{table}[]
	\centering
	\begin{tabular}{l|l}
		\hline
		\\ [-0.9em]
		\multicolumn{1}{c|}{Top-level} & \multicolumn{1}{c}{Low-level} \\ \hline
		Sport                & Hockey, Baseball, Autos, Motorcycles \\ \hline
		Politics                & Guns, Mideast, Misc \\ \hline
		Religion                & Christian, Atheism, Misc \\
		\hline
	\end{tabular}
	\caption{20NG category mappings}				
	\label {category-mappings}
\end{table}

\begin{table*}[]
	\centering
	\begin{tabular}{c|cc|cc|cc} 
		\hline
		&&&& \\ [-0.7em]
		\multicolumn{1}{c|}{Method}                   & \multicolumn{2}{c|}{Hockey x Baseball} & \multicolumn{2}{c|}{Autos x Motorcycles} & \multicolumn{2}{c}{Guns x Mideast x Misc}  \\ [0.3em]
		\hline
		&&&&&&\\ [-0.9em]
		ESA                      & 94.60 & @425                       & 72.70&@325                        & 70.00&@500  \\ [0.1em] \hline 
		&&&&&&\\ [-0.9em]
		CCX (equal)                       & 94.60&@20                       & -&-                       & 70.33&@60 \\ [0.1em]
		CRX (equal)                      & 94.60&@60              & 73.10&@4               & 70.00&@7 \\ [0.1em] \hline
		&&&&\\ [-0.9em]
		WE$_{max}$                     & 86.85&@65                       & 76.15&@375                       & 72.20&@300 \\ [0.1em]
		WE$_{hung}$                      & 95.20&@325              & 73.75&@300               & 71.70&@275 
		\\ [0.1em] \hline
		&&&&\\ [-0.9em]
		CCX (best)                      & 95.10&@125                       & 69.70&@7                       & 72.47&@250 \\ [0.1em]
		+bootstrap                      & 95.90&@450                       & 74.25&@12                       & \textbf{77.43}&@5 \\ \hline 
		&&&&&&\\ [-0.9em] 
		CRX (best)                      & 95.65&@425              & \textbf{79.20}&@14               & 73.40&@70 \\ 
		+bootstrap                      & \textbf{95.90}&@350              & 73.25&@12               & 77.03&@10 \\ 
		\hline
	\end{tabular}
	\caption{Evaluation results of dataless document classification of fine-grained classes measured in micro-averaged F1 along with \# of dimensions (concepts) in the BoC at which corresponding performance is achieved. 
	}		
	\label{20ng-fine}
\end{table*}

\subsubsection{Compared Systems}
We compare our models to three previous methods:
\begin{enumerate}[topsep=0pt]
	\itemsep0em
	\item \textbf{\emph{ESA}} which computes the cosine similarity between target labels and instance documents using the sparse BoC vectors.
	\item \textbf{\emph{WE}$_{max}$} \& \textbf{\emph{WE}$_{hung}$} which were proposed by \shortciteA{songunsupervised} for BoC densification using embeddings obtained from Word2Vec. As the authors reported, we fix the minimum similarity threshold to 0.85. WE$_{max}$ finds the best match for each concept, while WE$_{hung}$ utilizes the Hungarian algorithm to find the best concept-concept alignment on one-to-one basis. Both mechanisms have polynomial degree time complexity.
\end{enumerate}	

\subsubsection{Results}
Table \ref{20ng-fine} presents the results of fine-grained dataless classification measured in micro-averaged F1. As we can notice, ESA achieves its peak performance with a few hundred dimensions of the sparse BoC vector. Using our densification mechanism, both the CRX \& CCX models achieve equal performance to ESA at many fewer dimensions. Densification using the CRX model embeddings gives the best F1 scores on the three tasks. Interestingly, the CRX model improves the F1 score by $\sim$7\% using only 14 concepts on \emph{Autos} vs. \emph{Motorcycles}, and by $\sim$3\% using 70 concepts on \emph{Guns} vs. \emph{Mideast} vs. \emph{Misc}. The CCX model, still performs better than ESA on 2 out of the 3 tasks. Both WE$_{max}$ and WE$_{hung}$ improve the performance over ESA but not as our CRX model.

\begin{figure*}[t!]
	\centering
	\scriptsize
	\begin{tabular}{@{}ccc@{}}
		\includegraphics[width=.5\textwidth]{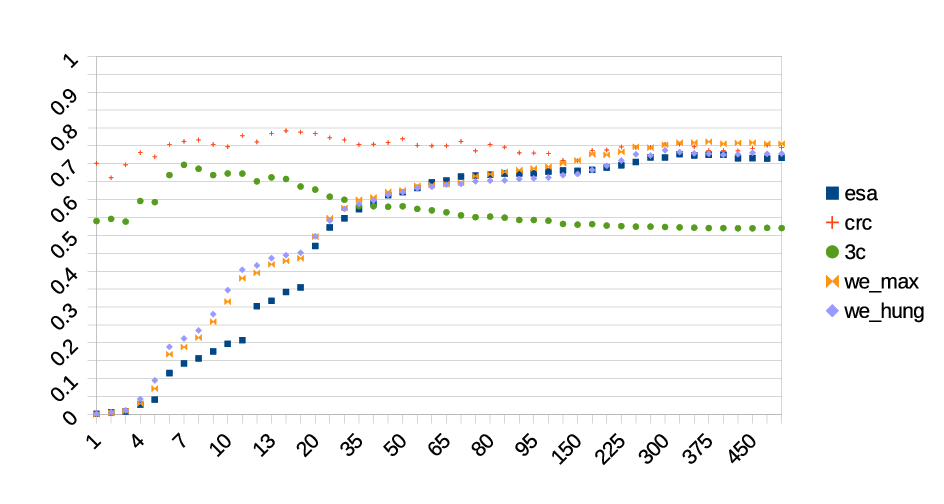} &
		\includegraphics[width=.5\textwidth]{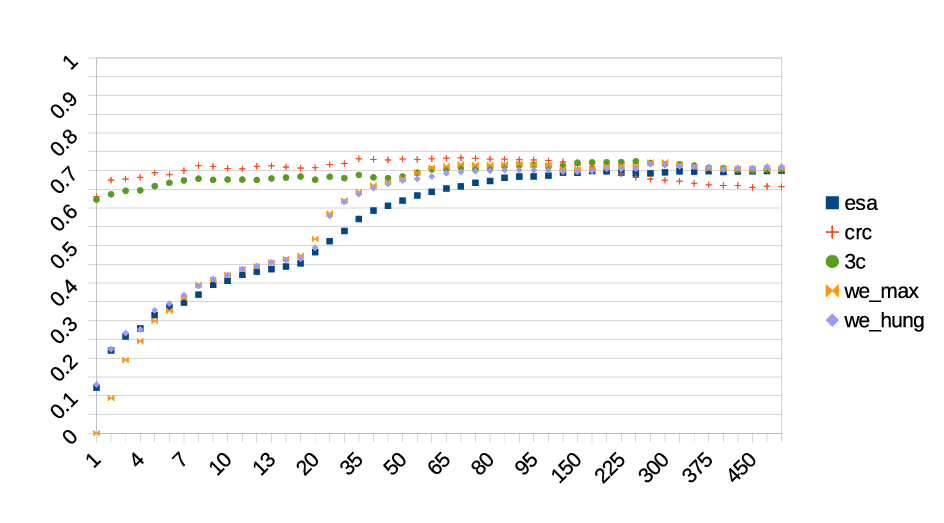} \\  [-1.4em]
		Autos vs. Motors & Guns vs. Mideast vs. Misc \\
		\multicolumn{2}{c}{\includegraphics[width=.5\textwidth]{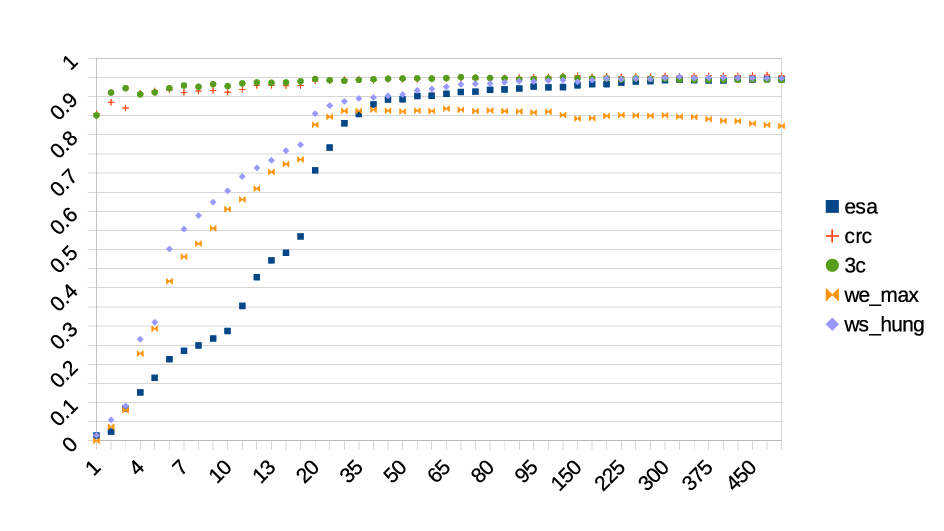}}
		\\ [-1.4em]
		\multicolumn{2}{c}{Hockey vs. Baseball}
	\end{tabular}
	\caption{micro-averaged F1 scores of fine-grained classes when varying the \# of BoC dimensions.}
	\label{20ng-fine-fig}
\end{figure*}

When we applied bootstrapping, the performance of the CCX model improved slightly on \emph{Hockey} vs. \emph{Baseball}, but significantly ($\sim$5\%) on the other two tasks achieving best performance on the third task with just 5 concepts. Bootstrapping with the CRX model has a similar effect to the CCX model except for \emph{Autos} vs. \emph{Motorcycles} where performance degraded significantly. To better understand this behavior, we analyzed the results as bootstrapping progresses at 14 concepts like CRX (best). We noticed that, at the very early iterations of Algorithm \ref{alg1}, many instances belonging to \emph{Autos} were closer to \emph{Motorcycles} with similarity scores between 0.90-0.95. And when using those instances to bootstrap \emph{Motorcycles}, they caused \emph{topic drift} moving \emph{Motorcycles}'s centroid toward \emph{Autos}, and eventually causing relatively lower accuracy scores. 

In order to better illustrate the robustness of our densification mechanism when varying the number of BoC dimensions, we measured F1 scores of each task as a function of the number of BoC dimensions used for densification. As we see in Figure \ref{20ng-fine-fig}, with \emph{one} concept we can achieve high F1 scores compared to ESA which achieves \emph{zero} or very low score. Moreover, \emph{near-peak performance} is achievable with the top 50 or less dimensions. We can also notice that, as we increase the number of dimensions, both WE$_{max}$ and WE$_{hung}$ densification methods have the same undesired monotonic pattern like ESA. Actually, the imposed threshold by these methods does not allow for full dense representation of the BoC vector and therefore at low dimensions we still see low overall F1 score. Our proposed densification mechanism besides its low cost, produces fully densified representations allowing good similarities at low dimensions.

\begin{figure*}[]
	\centering
	\scriptsize
	\begin{tabular}{@{}cc@{}}
		\includegraphics[width=.5\textwidth]{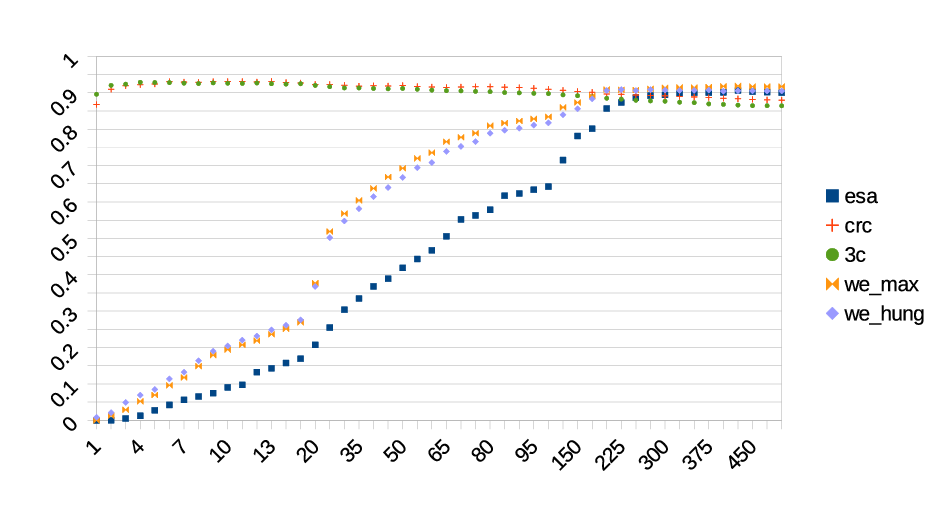} &
		\includegraphics[width=.5\textwidth]{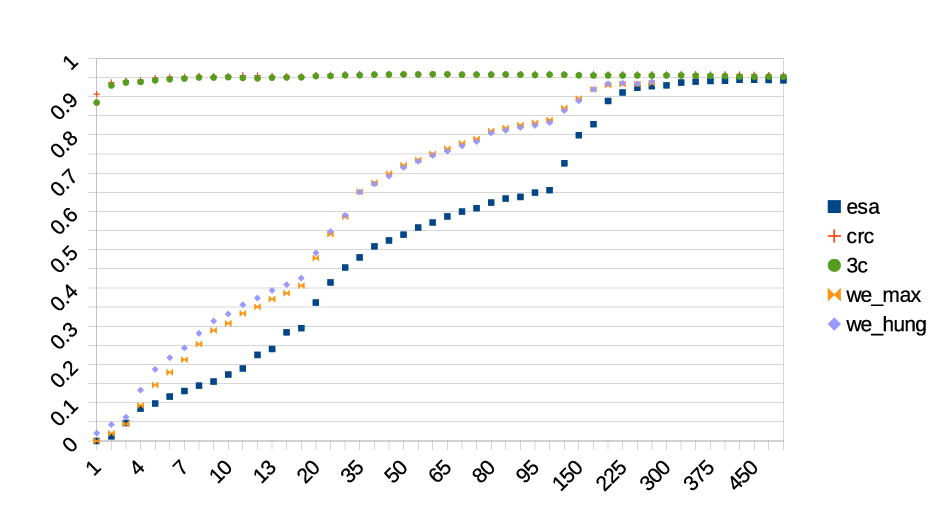} \\ [-1.4em]
		Sport vs. Politics & Sport vs. Religion \\
	\end{tabular}
	\caption{micro-averaged F1 scores of coarse-grained classes when varying the  \# of concepts (dimensions) in the BoC from 1 to 500.}
	\label{20ng-coarse-fig}
\end{figure*}

\begin{table}[]
	\centering		
	\begin{tabular}{c|cc|cc} 
		\hline
		&&&& \\ [-0.9em]
		\multicolumn{1}{c|}{Method}                   & \multicolumn{2}{c|}{Sport x Politics} & \multicolumn{2}{c}{Sport x Religion}           \\ [0.3em]
		\hline
		&&&&\\ [-0.9em]
		ESA  & 90.63&@425 & 94.39&@450          \\ [0.1em] \hline 
		&&&&\\ [-0.9em]
		CCX (equal) & 92.04&@2 & 95.11&@6          \\ [0.1em]
		CRX (equal) & 90.99&@2 & 94.81&@5           \\ 
		[0.1em] \hline
		&&&&\\ [-0.9em]
		WE$_{max}$                     & 91.89&@425                       &  93.99&@425 \\ [0.1em]
		WE$_{hung}$                      & 90.89&@275              & 94.16&@450 
		\\ \hline
		&&&&\\ [-0.9em]
		CCX (best) & 92.89&@4 & 95.86&@60          \\ 
		+bootstrap & \textbf{93.20}&@10 & 95.13&@225          \\ \hline
		&&&&\\ [-0.9em]
		CRX (best)  & 93.12&@13  & \textbf{95.91}&@95 \\ 
		+bootstrap  & 92.96&@13  & 95.53&@70 \\ 
		\hline
	\end{tabular}
	\caption{Evaluation results of dataless document classification of coarse-grained classes measured in micro-averaged F1 along with \# of dimensions (concepts) at which corresponding performance is achieved. 
	}
	\label{20ng-coarse}
\end{table}

Results of coarse-grained classification are presented in Table \ref{20ng-coarse}. Classification at the top level is easier than the fine-grained level. Nevertheless, as with fine-grained classification, ESA still peaks with a few hundred dimensions of the sparse BoC vector. 
Both the CRX \& CCX models achieve equal performance to ESA at very few dimensions ($\le\!6$). Densification using the CRX model embeddings still performs the best on both tasks. Interestingly, the CCX model gives very close F1 scores to the CRX model at less dimensions (@4 with \emph{Sport} vs. \emph{Politics}, and @60 with \emph{Sport} vs. \emph{Religion}) indicating its competitive advantage when training computational cost is a decisive criteria. The CCX model, still performs better than ESA, WE$_{max}$, and WE$_{hung}$ on both tasks.

Bootstrapping did not improve the results on this task significantly (if any). As we can notice in Table \ref{20ng-coarse}, the accuracy without bootstrapping is already high indicating that the initial prototype vector (centroid) of each class is representative enough of the instances to be classified.

Figure \ref{20ng-coarse-fig} shows F1 scores of coarse-grained classification when varying the \# of BoC dimensions used for densification. The same pattern of achieving \emph{near-peak performance} at very few dimensions recur with the CRX \& CCX models. ESA using the sparse BoC vectors achieves low F1 up until few hundred dimensions are considered. Even with the costly  WE$_{max}$ and WE$_{hung}$ densifications, performance sometimes decreases.

\begin{table*}[]
	\centering
	\begin{tabular}{c|cc|cc|cc} 
		\hline
		&&&& \\ [-0.7em]
		\multicolumn{1}{c|}{Method}                   & \multicolumn{2}{c|}{Hockey x Baseball} & \multicolumn{2}{c|}{Autos x Motorcycles} & \multicolumn{2}{c}{Guns x Mideast x Misc}  \\ [0.3em]
		\hline
		&&&&&&\\ [-0.9em]
		SVM                      & 91.61 & @85\%                       & 79.25&@20\%                        & 77.56&@25\%  \\ [0.1em] \hline 
		&&&&&&\\ [-0.9em]
		CCX (best)                       & 95.90&@450                       & 74.25&@12                       & 77.43&@5 \\ [0.1em]
		CRX (best)                      & 95.90&@350              & 79.20&@14               & 77.03&@10 \\ [0.1em] \hline
		
	\end{tabular}
	\caption{Evaluation results of dataless document classification of coarse-grained classes vs. supervised classification with SVM measured in micro-averaged F1 along with \# of dimensions (concepts) for CRX \& CCX or \% of labeld samples for SVM at which corresponding performance is achieved.}		
	\label{20ng-fine-supervised}
\end{table*}

\subsubsection{Dataless vs. Supervised Classification}
We performed a pilot experiment to demonstrate the value of the dataless classification scheme in the absence or difficulty of obtaining labeled data for training a supervised classifier. For this purpose, we used a Support Vector Machine (SVM) classifier  with a linear kernel, leveraging the scikit-learn machine learning library \shortcite{pedregosa2011scikit} to perform classification of fine-grained classes (cf. Table \ref{category-mappings}). We trained the SVM classifier with labeled samples ranging between 10\% to 90\% of the total number of samples for each task and evaluate the performance on the rest. The results in Table \ref{20ng-fine-supervised} shows the \% of labeled sampled needed for training SVM in order to achieve equal performance to dataless classification with CRX and CCX. As we can notice, with \emph{Hockey} vs \emph{Baseball}, the SVM classifier can't reach the same performance as either models and peaks when trained on 85\% ($\sim$1700 samples) of the data. With the \emph{Autos} vs \emph{Motorcycles} and \emph{Guns} vs \emph{Mideast} vs \emph{Misc}, the SVM classifier achieves equal performance when trained on 20\% ($\sim$400 samples) and 25\% ($\sim$750 samples) of the data respectively. These results demonstrate the competitiveness of our models to supervised classification even when training data is available. And its superiority when training data is scarce.

\begin{table*}[]
	\centering
	\footnotesize
	\begin{tabular}{|c|l|l|}
		\hline
		\textbf{Concept} & \multicolumn{1}{c|}{\textbf{Concept Raw Context (CRX)}}                                                     & \multicolumn{1}{c|}{\textbf{Concept-Concept Context (CCX)}} \\ \hline
		&& \\ [-0.9em]
		YouTube                           & \begin{tabular}[c]{@{}l@{}}Vevo\\ Facebook\\ SoundCloud\\ Vimeo\\ Viral video\end{tabular}    & \begin{tabular}[c]{@{}l@{}}Viral video\\ Vimeo\\ Vevo\\ Video blog\\ Dailymotion\end{tabular}     \\ \hline
		&& \\ [-0.9em]
		Harvard University                &    \begin{tabular}[c]{@{}l@{}}Yale University\\ Princeton University\\ Brown University\\ Columbia University\\ Boston University\end{tabular}                                               & \begin{tabular}[c]{@{}l@{}}Harvard Kennedy School\\ Cambridge, Massachusetts\\ Harvard College\\ Radcliffe College \\ Harvard Society of Fellows\end{tabular}                                                 \\ \hline
		&& \\ [-0.9em]
		Black hole                        & \begin{tabular}[c]{@{}l@{}}Neutron star\\ Accretion disk\\ Primordial black hole\\ Supermassive black hole\\ Event horizon\end{tabular} & \begin{tabular}[c]{@{}l@{}}Event horizon\\ Neutron star\\ Gravitational singularity\\ Wormhole\\ Hawking radiation\end{tabular}                            \\ \hline
		&& \\ [-0.9em]
		\begin{tabular}[c]{@{}c@{}}X-Men: \\Days of Future Past\end{tabular}                              & \begin{tabular}[c]{@{}l@{}}X-Men: Apocalypse\\ X-Men: First Class\\ Deadpool (film)\\ Avengers: Age of Ultron\\ X-Men: The Last Stand\end{tabular}                                   & \begin{tabular}[c]{@{}l@{}}X-Men: Apocalypse\\ The Wolverine (film)\\ X-Men: First Class\\ John Paesano\\ William Stryker\end{tabular}                               \\ \hline
	\end{tabular}
	\caption{Top-5 related concepts from CRX \& CCX models for sample target concepts}
	\label{qualitative}	
\end{table*}

\section{Discussion \& Conclusion}
\label{conclusion}
In this paper we proposed two models for learning neural embeddings of explicit concepts based on the skip-gram model. Explicit concepts are lexical expressions (single or multiwords) that denote an idea, event, or an object and typically have a set of properties associated with it. In the models presented here, our concept space is the set of all \emph{Wikipedia} article titles. We proposed learning concept representations from concept mentions/references in \emph{Wikipedia} making our models applicable to other open domain and domain specific free-text corpora by firstly wikifying\footnote{Wikification is the process of identifying mentions of concepts and entities in a given free-text and linking them to \emph{Wikipedia}} the text and then learning from concept mentions.

It is clear from the presented results that, the CRX model outperforms the CCX model on tasks that require topical coherence among the concepts vectors (e.g. concept categorization), while the CCX model is advantageous in tasks that require topical relatedness (e.g., measuring entity relatedness). To better show this difference qualitatively, we present a qualitative analysis of both models in Table \ref{qualitative} (target concepts are similar to those reported by \shortciteA{hu2015entity}).

As we can notice, the CRX model tends to emphasize concept \emph{topical and categorical similarity}, while the CCX model tends to more emphasize \emph{concept relatedness}. For example, the top-5 concepts closest to \emph{"Harvard University"} using CRX are all universities. While, the CCX model top-5 concepts include, besides educational institutions, location (\emph{"Cambridge, Massachusetts"}) and an affiliated group (\emph{"Harvard Society of Fellows"}). The same pattern can be noticed with the \emph{"X-Men"} movie where we get similar genre movies with CRX. While we get related characters such as \emph{"William Stryker"}\footnote{https://en.wikipedia.org/wiki/William\_Stryker} with CCX. 

Based on these observations, we claim that the CCX model would be beneficial in situations where \emph{diversity} is more desired than \emph{topical coherence}. This claim is also supported by the results we obtained on the concept categorization and dataless densification tasks. On concept categorization, the performance gap between CRX and CCX was large with almost all datasets. On dataless classification, the performance gap was large with documents belonging topics with nuance differences (i.e., \emph{Autos} vs. \emph{Motorcycles}), but with other classes which have clear distinctions, the CCX performance was very competitive to CRX (e.g., \emph{Hockey} vs \emph{Baseball}).

In this paper, we also proposed an \emph{efficient} and \emph{effective} mechanism for BoC densification which outperformed the previously proposed densification schemes on dataless document classification. Unlike these previous densification mechanisms, our method \emph{scales linearly} with the number of the BoC dimensions. In addition, we demonstrated through the results how this efficient mechanism allows generating high quality dense BoC from few concepts alleviating the need of obtaining hundreds of concepts when generating the BoC in the first place.

Our learning method does not require training on a hierarchical concept category graph and is not tightly coupled to linked knowledge base. Rather, we learn concept representations using mentions in free-text corpora with annotated concept mentions which even if not available could be obtained through state-of-the-art entity linking systems.

Finally, the work presented in this paper serves two of our objectives: 1) it demonstrates utilizing textual knowledge bases to learn robust concept embeddings and hence increasing the {\it effectiveness} of the BoC representation to better capture semantic similarities between textual structures, and 2) it demonstrates utilizing the learned distributed concept vectors to increase the {\it efficiency} of the semantic representations in terms of space and computational complexities.


\vskip 0.2in
\bibliography{ConceptEmbeddings}

\begin{thebibliography}{}

\bibitem[\protect\BCAY{Baroni\ \BBA\ Lenci}{Baroni\ \BBA\
  Lenci}{2010}]{baroni2010distributional}
Baroni, M.\BBACOMMA\  \BBA\ Lenci, A. \BBOP2010\BBCP.
\newblock \BBOQ Distributional memory: A general framework for corpus-based
  semantics\BBCQ\
\newblock {\Bem Computational Linguistics}, {\Bem 36\/}(4), 673--721.

\bibitem[\protect\BCAY{Bordes, Usunier, Garcia-Duran, Weston,\ \BBA\
  Yakhnenko}{Bordes et~al.}{2013}]{bordes2013translating}
Bordes, A., Usunier, N., Garcia-Duran, A., Weston, J., \BBA\ Yakhnenko, O.
  \BBOP2013\BBCP.
\newblock \BBOQ Translating embeddings for modeling multi-relational data\BBCQ\
\newblock In {\Bem Advances in neural information processing systems}, \BPGS\
  2787--2795.

\bibitem[\protect\BCAY{Ceccarelli, Lucchese, Orlando, Perego,\ \BBA\
  Trani}{Ceccarelli et~al.}{2013}]{ceccarelli2013learning}
Ceccarelli, D., Lucchese, C., Orlando, S., Perego, R., \BBA\ Trani, S.
  \BBOP2013\BBCP.
\newblock \BBOQ Learning relatedness measures for entity linking\BBCQ\
\newblock In {\Bem Proceedings of the 22nd ACM international conference on
  Information \& Knowledge Management}, \BPGS\ 139--148. ACM.

\bibitem[\protect\BCAY{Chang, Ratinov, Roth,\ \BBA\ Srikumar}{Chang
  et~al.}{2008}]{chang2008importance}
Chang, M.-W., Ratinov, L.-A., Roth, D., \BBA\ Srikumar, V. \BBOP2008\BBCP.
\newblock \BBOQ Importance of semantic representation: Dataless
  classification.\BBCQ\
\newblock In {\Bem AAAI}, \lowercase{\BVOL}~2, \BPGS\ 830--835.

\bibitem[\protect\BCAY{Collobert, Weston, Bottou, Karlen, Kavukcuoglu,\ \BBA\
  Kuksa}{Collobert et~al.}{2011}]{collobert2011natural}
Collobert, R., Weston, J., Bottou, L., Karlen, M., Kavukcuoglu, K., \BBA\
  Kuksa, P. \BBOP2011\BBCP.
\newblock \BBOQ Natural language processing (almost) from scratch\BBCQ\
\newblock {\Bem Journal of Machine Learning Research}, {\Bem 12\/}(Aug),
  2493--2537.

\bibitem[\protect\BCAY{Gabrilovich\ \BBA\ Markovitch}{Gabrilovich\ \BBA\
  Markovitch}{2007}]{gabrilovich2007computing}
Gabrilovich, E.\BBACOMMA\  \BBA\ Markovitch, S. \BBOP2007\BBCP.
\newblock \BBOQ Computing semantic relatedness using wikipedia-based explicit
  semantic analysis.\BBCQ\
\newblock In {\Bem IJCAI}, \lowercase{\BVOL}~7, \BPGS\ 1606--1611.

\bibitem[\protect\BCAY{Harris}{Harris}{1954}]{harris1954distributional}
Harris, Z.~S. \BBOP1954\BBCP.
\newblock \BBOQ Distributional structure.\BBCQ\
\newblock {\Bem Word}.

\bibitem[\protect\BCAY{Hoffart, Seufert, Nguyen, Theobald,\ \BBA\
  Weikum}{Hoffart et~al.}{2012}]{hoffart2012kore}
Hoffart, J., Seufert, S., Nguyen, D.~B., Theobald, M., \BBA\ Weikum, G.
  \BBOP2012\BBCP.
\newblock \BBOQ Kore: keyphrase overlap relatedness for entity
  disambiguation\BBCQ\
\newblock In {\Bem Proceedings of the 21st ACM international conference on
  Information and knowledge management}, \BPGS\ 545--554. ACM.

\bibitem[\protect\BCAY{Hu, Huang, Deng, Gao,\ \BBA\ Xing}{Hu
  et~al.}{2015}]{hu2015entity}
Hu, Z., Huang, P., Deng, Y., Gao, Y., \BBA\ Xing, E.~P. \BBOP2015\BBCP.
\newblock \BBOQ Entity hierarchy embedding\BBCQ\
\newblock In {\Bem Proceedings of The 53rd Annual Meeting of the Association
  for Computational Linguistics}.

\bibitem[\protect\BCAY{Hua, Wang, Wang, Zheng,\ \BBA\ Zhou}{Hua
  et~al.}{2015}]{hua2015short}
Hua, W., Wang, Z., Wang, H., Zheng, K., \BBA\ Zhou, X. \BBOP2015\BBCP.
\newblock \BBOQ Short text understanding through lexical-semantic
  analysis\BBCQ\
\newblock In {\Bem Data Engineering (ICDE), 2015 IEEE 31st International
  Conference on}, \BPGS\ 495--506. IEEE.

\bibitem[\protect\BCAY{Huang, Heck,\ \BBA\ Ji}{Huang
  et~al.}{2015}]{huang2015leveraging}
Huang, H., Heck, L., \BBA\ Ji, H. \BBOP2015\BBCP.
\newblock \BBOQ Leveraging deep neural networks and knowledge graphs for entity
  disambiguation\BBCQ\
\newblock {\Bem arXiv preprint arXiv:1504.07678}.

\bibitem[\protect\BCAY{Hulpus, Prangnawarat,\ \BBA\ Hayes}{Hulpus
  et~al.}{2015}]{hulpucs2015path}
Hulpus, I., Prangnawarat, N., \BBA\ Hayes, C. \BBOP2015\BBCP.
\newblock \BBOQ Path-based semantic relatedness on linked data and its use to
  word and entity disambiguation\BBCQ\
\newblock In {\Bem International Semantic Web Conference}, \BPGS\ 442--457.
  Springer.

\bibitem[\protect\BCAY{Kim, Wang,\ \BBA\ Oh}{Kim et~al.}{2013}]{kim2013context}
Kim, D., Wang, H., \BBA\ Oh, A.~H. \BBOP2013\BBCP.
\newblock \BBOQ Context-dependent conceptualization.\BBCQ\
\newblock In {\Bem IJCAI}, \BPGS\ 2330--2336.

\bibitem[\protect\BCAY{Lang}{Lang}{1995}]{lang1995newsweeder}
Lang, K. \BBOP1995\BBCP.
\newblock \BBOQ Newsweeder: Learning to filter netnews\BBCQ\
\newblock In {\Bem Proceedings of the 12th international conference on machine
  learning}, \BPGS\ 331--339.

\bibitem[\protect\BCAY{Li, Wang, Zhu, Wang,\ \BBA\ Wu}{Li
  et~al.}{2013}]{li2013computing}
Li, P., Wang, H., Zhu, K.~Q., Wang, Z., \BBA\ Wu, X. \BBOP2013\BBCP.
\newblock \BBOQ Computing term similarity by large probabilistic isa
  knowledge\BBCQ\
\newblock In {\Bem Proceedings of the 22nd ACM international conference on
  Conference on information \& knowledge management}, \BPGS\ 1401--1410. ACM.

\bibitem[\protect\BCAY{Li, Zheng, Tian, Hu, Iyer,\ \BBA\ Sycara}{Li
  et~al.}{2016}]{li2016joint}
Li, Y., Zheng, R., Tian, T., Hu, Z., Iyer, R., \BBA\ Sycara, K. \BBOP2016\BBCP.
\newblock \BBOQ Joint embedding of hierarchical categories and entities for
  concept categorization and dataless classification\BBCQ\
\newblock {\Bem arXiv preprint arXiv:1607.07956}.

\bibitem[\protect\BCAY{Mikolov, Chen, Corrado,\ \BBA\ Dean}{Mikolov
  et~al.}{2013a}]{mikolov2013efficient}
Mikolov, T., Chen, K., Corrado, G., \BBA\ Dean, J. \BBOP2013a\BBCP.
\newblock \BBOQ Efficient estimation of word representations in vector
  space\BBCQ\
\newblock {\Bem arXiv preprint arXiv:1301.3781}.

\bibitem[\protect\BCAY{Mikolov, Sutskever, Chen, Corrado,\ \BBA\ Dean}{Mikolov
  et~al.}{2013b}]{mikolov2013distributed}
Mikolov, T., Sutskever, I., Chen, K., Corrado, G.~S., \BBA\ Dean, J.
  \BBOP2013b\BBCP.
\newblock \BBOQ Distributed representations of words and phrases and their
  compositionality\BBCQ\
\newblock In {\Bem Advances in neural information processing systems}, \BPGS\
  3111--3119.

\bibitem[\protect\BCAY{Milne\ \BBA\ Witten}{Milne\ \BBA\
  Witten}{2008}]{milne2008learning}
Milne, D.\BBACOMMA\  \BBA\ Witten, I.~H. \BBOP2008\BBCP.
\newblock \BBOQ Learning to link with wikipedia\BBCQ\
\newblock In {\Bem Proceedings of the 17th ACM conference on Information and
  knowledge management}, \BPGS\ 509--518. ACM.

\bibitem[\protect\BCAY{Papadimitriou\ \BBA\ Steiglitz}{Papadimitriou\ \BBA\
  Steiglitz}{1982}]{papadimitriou1982combinatorial}
Papadimitriou, C.~H.\BBACOMMA\  \BBA\ Steiglitz, K. \BBOP1982\BBCP.
\newblock {\Bem Combinatorial optimization: algorithms and complexity}.
\newblock Courier Corporation.

\bibitem[\protect\BCAY{Pedregosa, Varoquaux, Gramfort, Michel, Thirion, Grisel,
  Blondel, Prettenhofer, Weiss, Dubourg, et~al.}{Pedregosa
  et~al.}{2011}]{pedregosa2011scikit}
Pedregosa, F., Varoquaux, G., Gramfort, A., Michel, V., Thirion, B., Grisel,
  O., Blondel, M., Prettenhofer, P., Weiss, R., Dubourg, V., et~al.
  \BBOP2011\BBCP.
\newblock \BBOQ Scikit-learn: Machine learning in python\BBCQ\
\newblock {\Bem Journal of machine learning research}, {\Bem 12\/}(Oct),
  2825--2830.

\bibitem[\protect\BCAY{Peng, Song,\ \BBA\ Roth}{Peng
  et~al.}{2016}]{peng2016event}
Peng, H., Song, Y., \BBA\ Roth, D. \BBOP2016\BBCP.
\newblock \BBOQ Event detection and co-reference with minimal supervision\BBCQ.
\newblock EMNLP.

\bibitem[\protect\BCAY{Pennington, Socher,\ \BBA\ Manning}{Pennington
  et~al.}{2014}]{pennington2014glove}
Pennington, J., Socher, R., \BBA\ Manning, C.~D. \BBOP2014\BBCP.
\newblock \BBOQ Glove: Global vectors for word representation\BBCQ\
\newblock {\Bem Proceedings of the Empiricial Methods in Natural Language
  Processing (EMNLP 2014)}, {\Bem 12}, 1532--1543.

\bibitem[\protect\BCAY{Rocchio}{Rocchio}{1971}]{rocchio1971relevance}
Rocchio, J.~J. \BBOP1971\BBCP.
\newblock
\newblock \BBOQ Relevance feedback in information retrieval\BBCQ.

\bibitem[\protect\BCAY{Schuhmacher\ \BBA\ Ponzetto}{Schuhmacher\ \BBA\
  Ponzetto}{2014}]{schuhmacher2014knowledge}
Schuhmacher, M.\BBACOMMA\  \BBA\ Ponzetto, S.~P. \BBOP2014\BBCP.
\newblock \BBOQ Knowledge-based graph document modeling\BBCQ\
\newblock In {\Bem Proceedings of the 7th ACM international conference on Web
  search and data mining}, \BPGS\ 543--552. ACM.

\bibitem[\protect\BCAY{Shalaby\ \BBA\ Zadrozny}{Shalaby\ \BBA\
  Zadrozny}{2015}]{shalaby2015measuring}
Shalaby, W.\BBACOMMA\  \BBA\ Zadrozny, W. \BBOP2015\BBCP.
\newblock \BBOQ Measuring semantic relatedness using mined semantic
  analysis\BBCQ\
\newblock {\Bem arXiv preprint arXiv:1512.03465}.

\bibitem[\protect\BCAY{Song\ \BBA\ Roth}{Song\ \BBA\
  Roth}{2014}]{song2014dataless}
Song, Y.\BBACOMMA\  \BBA\ Roth, D. \BBOP2014\BBCP.
\newblock \BBOQ On dataless hierarchical text classification.\BBCQ\
\newblock In {\Bem AAAI}, \BPGS\ 1579--1585.

\bibitem[\protect\BCAY{Song\ \BBA\ Roth}{Song\ \BBA\
  Roth}{2015}]{songunsupervised}
Song, Y.\BBACOMMA\  \BBA\ Roth, D. \BBOP2015\BBCP.
\newblock \BBOQ Unsupervised sparse vector densification for short text
  similarity\BBCQ\
\newblock In {\Bem Proceedings of NAACL}.

\bibitem[\protect\BCAY{Song, Wang, Wang, Li,\ \BBA\ Chen}{Song
  et~al.}{2011}]{song2011short}
Song, Y., Wang, H., Wang, Z., Li, H., \BBA\ Chen, W. \BBOP2011\BBCP.
\newblock \BBOQ Short text conceptualization using a probabilistic
  knowledgebase\BBCQ\
\newblock In {\Bem Proceedings of the Twenty-Second international joint
  conference on Artificial Intelligence-Volume Volume Three}, \BPGS\
  2330--2336. AAAI Press.

\bibitem[\protect\BCAY{Song, Wang,\ \BBA\ Wang}{Song
  et~al.}{2015}]{song2015open}
Song, Y., Wang, S., \BBA\ Wang, H. \BBOP2015\BBCP.
\newblock \BBOQ Open domain short text conceptualization: A generative+
  descriptive modeling approach.\BBCQ\
\newblock In {\Bem IJCAI}, \BPGS\ 3820--3826.

\bibitem[\protect\BCAY{Wang\ \BBA\ Wang}{Wang\ \BBA\
  Wang}{2016}]{understanding-short-texts}
Wang, Z.\BBACOMMA\  \BBA\ Wang, H. \BBOP2016\BBCP.
\newblock \BBOQ Understanding short texts\BBCQ\
\newblock In {\Bem the Association for Computational Linguistics (ACL)
  (Tutorial)}.

\bibitem[\protect\BCAY{Wang, Wang,\ \BBA\ Hu}{Wang et~al.}{2014}]{wang2014head}
Wang, Z., Wang, H., \BBA\ Hu, Z. \BBOP2014\BBCP.
\newblock \BBOQ Head, modifier, and constraint detection in short texts\BBCQ\
\newblock In {\Bem Data Engineering (ICDE), 2014 IEEE 30th International
  Conference on}, \BPGS\ 280--291. IEEE.

\bibitem[\protect\BCAY{Wang, Zhao, Wang, Meng,\ \BBA\ Wen}{Wang
  et~al.}{2015}]{wang2015query}
Wang, Z., Zhao, K., Wang, H., Meng, X., \BBA\ Wen, J.-R. \BBOP2015\BBCP.
\newblock
\newblock \BBOQ Query understanding through knowledge-based
  conceptualization\BBCQ.

\bibitem[\protect\BCAY{Witten\ \BBA\ Milne}{Witten\ \BBA\
  Milne}{2008}]{witten2008effective}
Witten, I.\BBACOMMA\  \BBA\ Milne, D. \BBOP2008\BBCP.
\newblock \BBOQ An effective, low-cost measure of semantic relatedness obtained
  from wikipedia links\BBCQ\
\newblock In {\Bem Proceeding of AAAI Workshop on Wikipedia and Artificial
  Intelligence: an Evolving Synergy, AAAI Press, Chicago, USA}, \BPGS\ 25--30.

\bibitem[\protect\BCAY{Wu, Li, Wang,\ \BBA\ Zhu}{Wu
  et~al.}{2012}]{wu2012probase}
Wu, W., Li, H., Wang, H., \BBA\ Zhu, K.~Q. \BBOP2012\BBCP.
\newblock \BBOQ Probase: A probabilistic taxonomy for text understanding\BBCQ\
\newblock In {\Bem Proceedings of the 2012 ACM SIGMOD International Conference
  on Management of Data}, \BPGS\ 481--492. ACM.

\bibitem[\protect\BCAY{Yamada, Shindo, Takeda,\ \BBA\ Takefuji}{Yamada
  et~al.}{2016}]{yamada2016joint}
Yamada, I., Shindo, H., Takeda, H., \BBA\ Takefuji, Y. \BBOP2016\BBCP.
\newblock \BBOQ Joint learning of the embedding of words and entities for named
  entity disambiguation\BBCQ\
\newblock {\Bem arXiv preprint arXiv:1601.01343}.

\bibitem[\protect\BCAY{Zwillinger\ \BBA\ Kokoska}{Zwillinger\ \BBA\
  Kokoska}{1999}]{zwillinger1999crc}
Zwillinger, D.\BBACOMMA\  \BBA\ Kokoska, S. \BBOP1999\BBCP.
\newblock {\Bem {CRC standard probability and statistics tables and formulae}}.
\newblock CRC.

\end{thebibliography}
\bibliographystyle{theapa}

\end{document}